\def\year{2020}\relax
\documentclass[letterpaper]{article} 
\usepackage{aaai20}  
\usepackage{times}  
\usepackage{helvet} 
\usepackage{courier}  
\usepackage[hyphens]{url}  
\usepackage{graphicx} 
\usepackage{graphicx}  
\usepackage{algorithm}
\usepackage{algorithmic}
\usepackage{amsmath}
\usepackage{amssymb}
\usepackage{multirow}
\usepackage{subcaption}

\title{Regularized Fine-grained Meta Face Anti-spoofing}
\author{Rui Shao, Xiangyuan Lan, Pong C. Yuen\\ 
	Department of Computer Science, Hong Kong Baptist University, Hong Kong\\
	\{ruishao, pcyuen\}@comp.hkbu.edu.hk, xiangyuanlan@life.hkbu.edu.hk 
}

\begin{document}

\maketitle

\begin{abstract}
Face presentation attacks have become an increasingly critical concern when face recognition is widely applied. Many face anti-spoofing methods have been proposed, but most of them ignore the generalization ability to unseen attacks. To overcome the limitation, this work casts face anti-spoofing as a domain generalization (DG) problem, and attempts to address this problem by developing a new meta-learning framework called Regularized Fine-grained Meta-learning. To let our face anti-spoofing model generalize well to unseen attacks, the proposed framework trains our model to perform well in the simulated domain shift scenarios, which is achieved by finding generalized learning directions in the meta-learning process. Specifically, the proposed framework incorporates the domain knowledge of face anti-spoofing as the regularization so that meta-learning is conducted in the feature space regularized by the supervision of domain knowledge. This enables our model more likely to find generalized learning directions with the regularized meta-learning for face anti-spoofing task. Besides, to further enhance the generalization ability of our model, the proposed framework adopts a fine-grained learning strategy that simultaneously conducts meta-learning in a variety of domain shift scenarios in each iteration. Extensive experiments on four public datasets validate the effectiveness of the proposed method. 
\end{abstract}

\section{Introduction}
Face recognition, as one of the computer vision techniques~\cite{Tracking_2019_TIE,Reid_2019_TIP}, has been successfully applied in a variety of applications in the real life, such as automated teller machines (ATMs), mobile payments, and entrance guard systems. Although much convenience is brought by the face recognition technique, many kinds of face presentation attacks (PA) also appear. Easy-accessible human faces from the Internet or social media can be abused to produce print attacks (i.e. based on the printed photo papers) or video replay attacks (i.e. based on the digital image/videos). These attacks can successfully hack a face recognition system deployed in a mobile phone or a laptop because those spoofs are visually extremely close to the genuine faces. Therefore, how to protect our face recognition systems against these presentation attacks has become an increasingly critical issue in the face recognition community.

\begin{figure}[t]
	\begin{center}
		\includegraphics[height=3.4cm, width=1\linewidth]{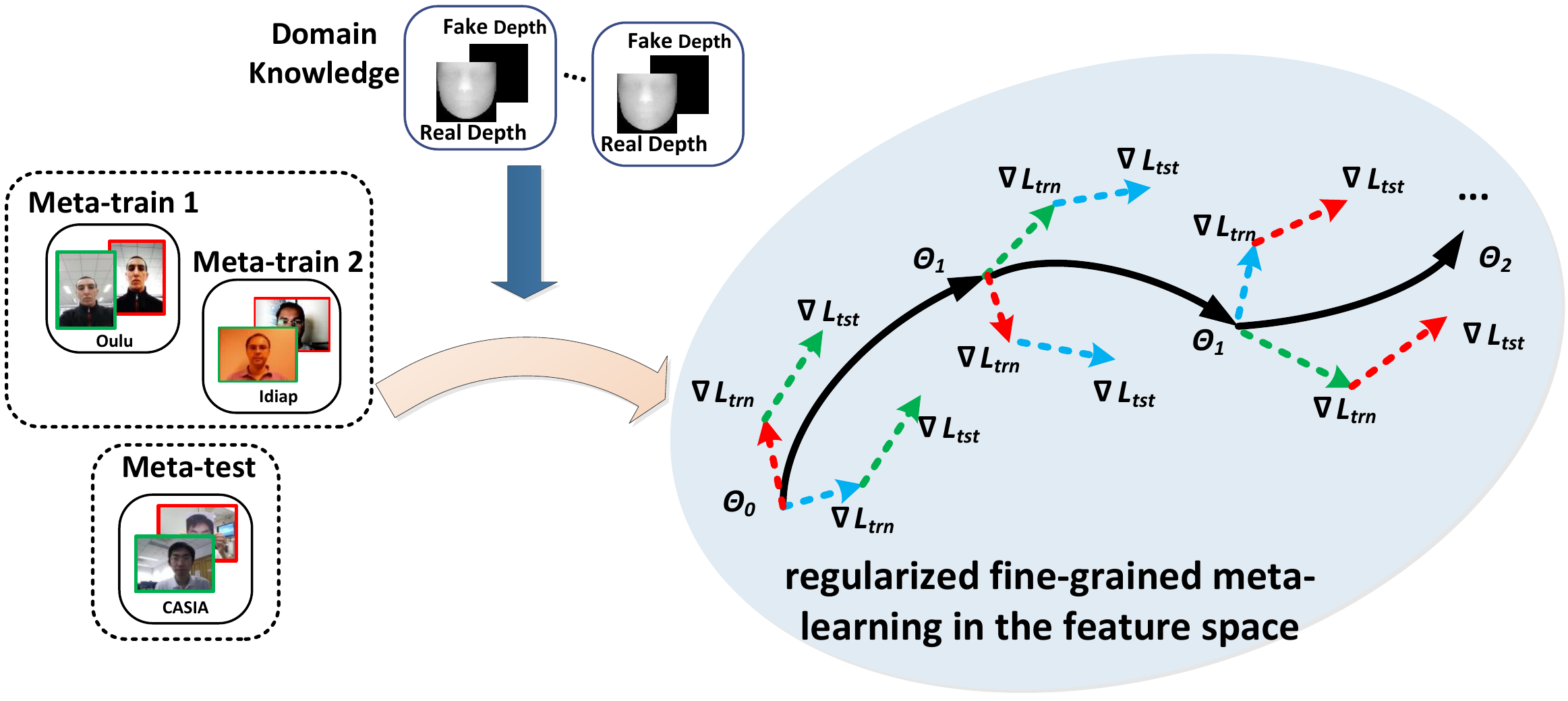}
	\end{center}
	\caption{Idea of the proposed regularized fine-grained meta-learning framework. By incorporating domain knowledge as regularization, meta-learning is conducted in the feature space regularized by the domain knowledge supervision. Thus, generalized learning directions are more likely to be found for task of face anti-spoofing. Besides, the proposed framework adopts a fine-grained learning strategy that simultaneously conducts meta-learning in a variety of domain shift scenarios. Thus, more abundant domain shift information of face anti-spoofing task can be exploited.}
\end{figure}

\begin{figure}[t]
	\begin{center}
		\includegraphics[height=4.5cm, width=1\linewidth]{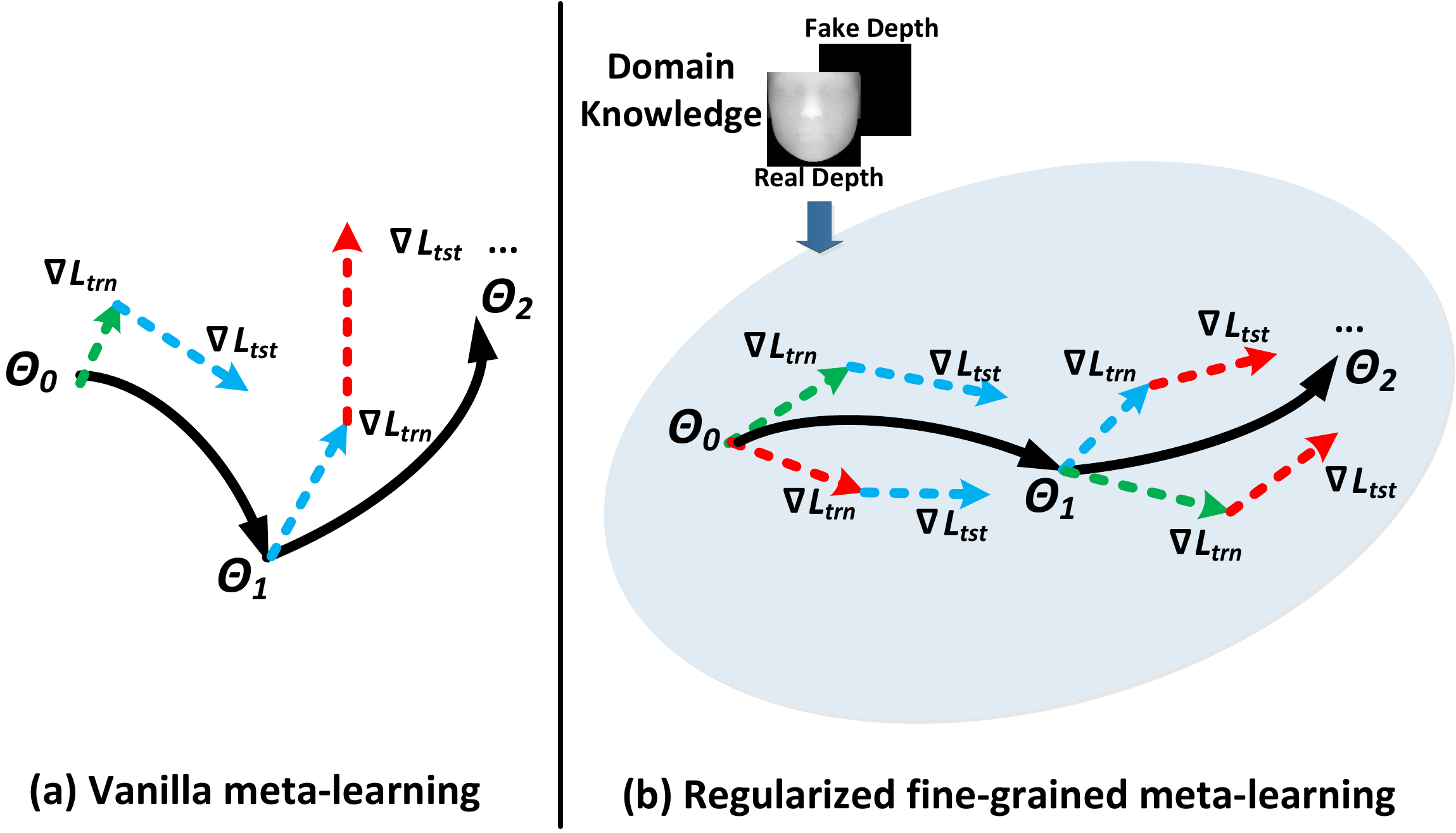}
	\end{center}
	\caption{Comparison of learning directions between (a) vanilla meta-learning, and (b) regularized fine-grained meta-learning. Three source domains are used as examples. Dotted arrows with different colors denote the learning directions (gradients) of meta-train ($\nabla L_{trn}$) and meta-test ($\nabla L_{tst}$) in different domains. Solid arrows denote the summarized learning directions of meta-optimization. $\theta_i$ $(i=0,1,...) $ are the updated model parameters in $i$-th iteration.}
\end{figure}

Many face anti-spoofing methods have been proposed. Appearance-based methods choose to extract various appearance cues to differentiate real and fake~\cite{2016TIFScolortxt,2016TIFScolortxt,2015TIFSida,2014arxivdeepfeat}; Temporal-based methods aim to do differentiation based on various temporal cues~\cite{2014EJIVPlbptop,2018TIFSdynamictext,RuiShao2018IJCB,2018ECCVrPPG,2018CVPRauxliary}. Although these methods obtain promising performance in intra-dataset experiments where training and testing data are from the same dataset, the performance dramatically degrades in cross-dataset experiments where models are trained on one dataset and tested on a related but shifted dataset. This is because existing face anti-spoofing methods capture the differentiation cues that are dataset biased~\cite{2011CVPRdatasetbias}, and thus cannot generalize well to unseen testing data that have different feature distribution compared to training data (mainly caused by different materials of attacks or recording environments).

To overcome this limitation, this paper casts face anti-spoofing as a domain generalization (DG) problem. Compared to the traditional unsupervised domain adaptation (UDA)~\cite{2018MMhada,Saito_2018_CVPR,Mancini_2018_CVPR,Zhang_2018_CVPR,Chen_2018_CVPR,Pinheiro_2018_CVPR,Tzeng_2017_CVPR,Bousmalis_2017_CVPR,Volpi_2018_CVPR,Zhangjing_2018_CVPR,Damodaran_2018_ECCV,Shao_2019_IETIP} that assume access to the labeled source domain data and unlabeled target domain data, DG assumes no access to target domain information. For DG, multiple source domains are exploited to learn the model which can generalize well to unseen test data in the target domain. For the task of face anti-spoofing, because we do not know what kind of attacks will be presented to our face recognition system, we have no clue on the testing dataset (target domain data) when we train our model so that DG is more suitable for our task.

Inspired by~\cite{Finn2017ICML,DaLi2017AAAI}, this paper aims to address problem of DG for face anti-spoofing in a \textit{meta-learning} framework. However, if we directly apply existing vanilla meta-learning for DG algorithms on the task of face anti-spoofing, the performance will be degraded due to the following two issues: 1) It is found that face anti-spoofing models only with binary class supervision discover arbitrary differentiation cues with poor generalization~\cite{2018CVPRauxliary}. As such, as illustrated in Fig. 2(a), if vanilla meta-learning algorithms are applied in face anti-spoofing only with the supervision of the binary class labels, the learning directions in the meta-train and meta-test steps will be arbitrary and biased, which makes it difficult for the meta-optimization step to summarize and find a generalized learning direction finally. 2) Vanilla meta-learning for DG methods~\cite{DaLi2017AAAI} coarsely divide multiple source domains into two groups to form one aggregated meta-train and one aggregated meta-test domains in each iteration of meta-learning. Thus only a single domain shift scenario is simulated in each iteration, which is sub-optimal for the task of face anti-spoofing. In order to equip the model with the generalization ability to unseen attacks of various scenarios, a variety of domain shift scenarios instead of a single one that are simulated for meta-learning is more optimal for the task of face anti-spoofing.

To address the above two issues, as illustrated in Fig. 1, this paper proposes a novel regularized fine-grained meta-learning framework. For the first issue, compared to binary class labels, domain knowledge specific to the task of face anti-spoofing can provide more generalized differentiation information. Therefore, as illustrated in Fig .2(b), the proposed framework incorporates the domain knowledge of face anti-spoofing as regularization into feature learning process so that meta-learning is conducted in the feature space regularized by the auxiliary supervision of domain knowledge. In this way, this regularized meta-learning can focus on more coordinated and better-generalized learning directions in the meta-train and meta-test for task of face anti-spoofing. Therefore, the summarized learning direction in the meta-optimization can guide face anti-spoofing model to exploit more generalized differentiation cues. Besides, for the second issue, the proposed framework adopts a fine-grained learning strategy as shown in Fig .2(b). This strategy divides source domains into multiple meta-train and meta-test domains, and jointly conducts meta-learning between each pair of them in each iteration. As such, a variety of domain shift scenarios are simultaneously simulated and thus more abundant domain shift information can be exploited in the meta-learning to train a generalized face anti-spoofing model.

\section{Related Work}

\textbf{Face Anti-spoofing Methods.} Current face anti-spoofing methods can be roughly categorized into appearance-based methods and temporal-based methods. Appearance-based methods are proposed to extract different appearance cues for attacks detection. Multi-scale LBP~\cite{2011IJCBmstexture} and color textures~\cite{2016TIFScolortxt} methods are proposed to extract various LBP descriptors in various color spaces for the differentiation between real/fake. Image distortion analysis~\cite{2015TIFSida} detects the surface distortions due to lower appearance quality of images or videos compared to the real face skin. Yang \emph{et al.}~\cite{2014arxivdeepfeat} trains CNN to extract discriminative deep features for real/fake faces classification. On the other hand, temporal-based methods aim to extract different temporal cues through multiple frames to differentiate real/fake faces. Dynamic texture methods proposed in~\cite{2014EJIVPlbptop,2018TIFSdynamictext,RuiShao2018IJCB} try to extract different facial motions. Liu \emph{et al.}~\cite{2016ECCVrPPG,2018ECCVrPPG} propose to capture discriminative rPPG signals from real/fake faces. ~\cite{2018CVPRauxliary} learns a CNN-RNN model to estimate the different face depth and rPPG signals between real/fake faces. However, the performance of both appearance and temporal-based methods become degraded in cross-datasets test where unseen attacks are encountered. This is because all the above methods are likely to extract some differentiation cues that are biased to specific materials of attacks or recording environments in training datasets. Comparatively, the proposed method conducts meta-learning for DG in the simulated domain shift scenarios, which is designed to make our model generalize well and capture more generalized differentiation cues for the task of face anti-spoofing. Note that a recent work~\cite{Shao_2019_CVPR} proposes multi-adversarial discriminative deep domain generalization for face anti-spoofing. It assumes that generalized differentiation cues can be discovered by searching a shared and discriminative feature space via adversarial learning. However, there is no guarantee that such a feature space exists among multiple source domains. Moreover, it needs to train multiple extra discriminators for all source domains. Comparatively, this paper does not need such a strong assumption and meta-learning can be conducted without training extra discriminators networks for adversarial learning, which is more efficient.

\noindent\textbf{Meta-learning for Domain Generalization Methods.} 

\noindent Unlike meta-learning for few-shot learning~\cite{Finn2017ICML}, meta-learning for DG is relatively less explored. MLDG~\cite{DaLi2017AAAI} designs a model-agnostic meta-learning for DG. Reptile~\cite{Reptile_2018_arXiv} is a general first-order meta-learning method that can be easily adapted into DG task. MetaReg~\cite{MetaReg_2018_NIPS} learns regularizers for DG in a meta-learning framework. However, directly applying the aforementioned methods in the task of face anti-spoofing may encounter the two issues mentioned above. Comparatively, our method conducts meta-learning in the feature space regularized by auxiliary supervision of domain knowledge within a fine-grained learning strategy. This contributes a more feasible meta-learning for DG in the task of face anti-spoofing.

\section{Proposed Method}

\begin{figure}[t]
	\begin{center}
		\includegraphics[height=4.3cm, width=1\linewidth]{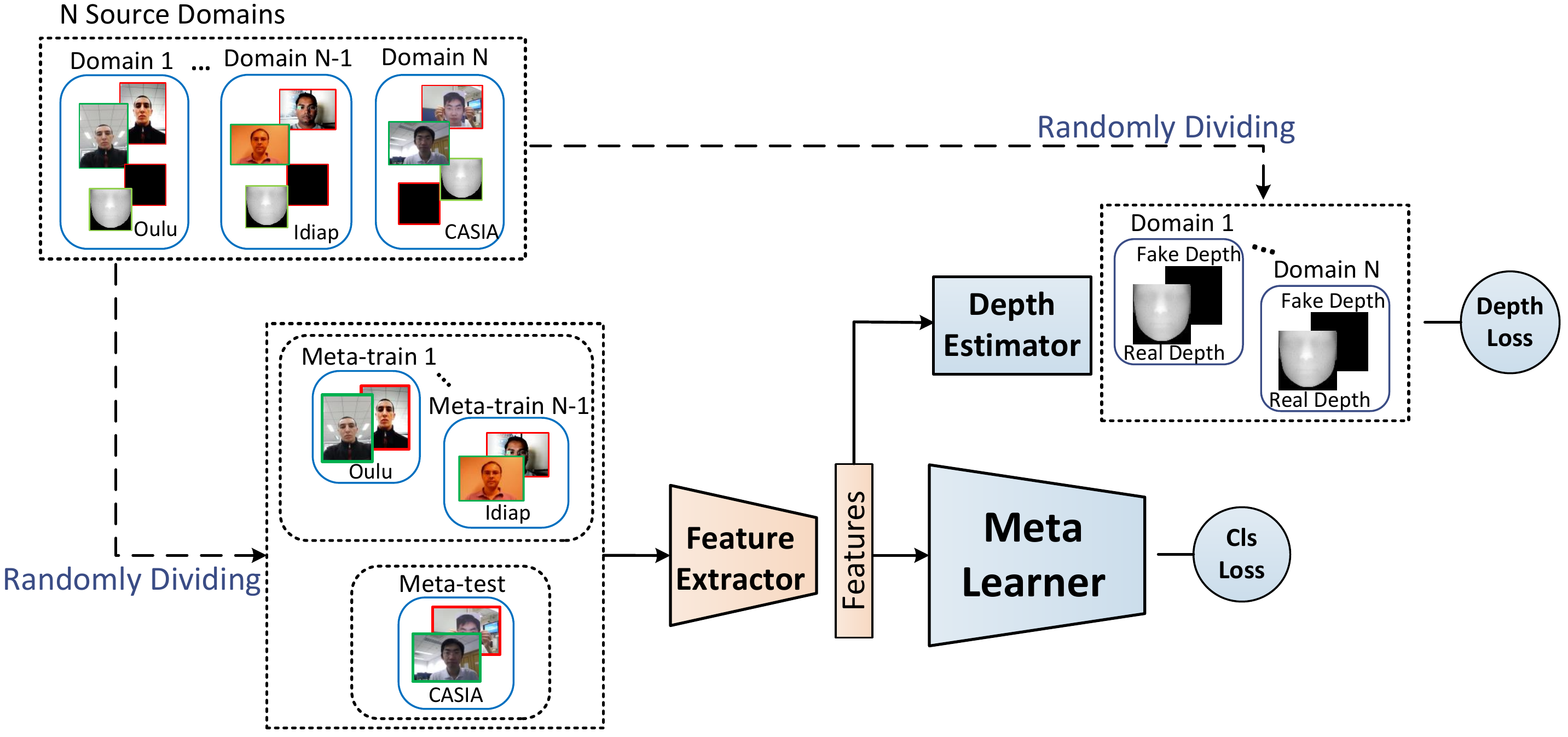}
	\end{center}
	\caption{Overview of proposed framework. We simulate domain shift by randomly dividing original $N$ source domains in each iteration. Supervision of domain knowledge is incorporated via depth estimator to regularize the learning process of feature extractor. Thus, meta learner conducts the meta-learning in the feature space regularized by the auxiliary supervision of domain knowledge. }
\end{figure}

The overall proposed framework is illustrated in Fig. 3.

\subsection{Domain Shift Simulating}

Suppose that we have access to $N$ source domains of face anti-spoofing task, denoted as $D=[D_1,D_2,...,D_N]$. The objective of DG for face anti-spoofing is to make the model trained on the $N$ source domains can generalize well to unseen attacks from the target domain. To this end, at each training iteration, we divide the original $N$ source domains by randomly selecting $N-1$ domains as meta-train domains (denoted as $D_{trn}$) and the remaining one as the meta-test domain (denoted as $D_{val}$). As such, the training and testing domain shift in the real world can be simulated. In this way, our model can learn how to perform well in the domain shift scenarios through many training iterations and thus learn to generalize well to unseen attacks. 

\subsection{Regularized Fine-grained Meta-learning}
Several existing vanilla meta-learning for DG methods can be applied to achieve the above objective. But their performance degrade for the task of face anti-spoofing due to the two issues mentioned in the introduction. To address these issues, this paper proposes a new meta-learning framework called regularized fine-grained meta-learning. In each meta-train and meta-test domain, we are provided with image and label pairs denoted as $x$ and $y$, where $y$ are ground truth with binary class labels ($y$ = 0/1 is the label of fake/real face). Compared to the binary class labels, domain knowledge specific to the face anti-spoofing task can provide more generalized differentiation information. This paper adopts the face depth map as the domain knowledge. By comparing the spatial information, it can be observed that live faces have face-like depth, while faces of attacks presented in the flat and planar papers or video screens have no face depth. In this way, for the first issue, we incorporate this domain knowledge as regularization into feature learning process so that meta-learning can be conducted in the feature space regularized by the auxiliary supervision of domain knowledge. Thus, this regularized meta-learning in the feature space can focus on better-generalized learning directions in meta-train and meta-test for task of face anti-spoofing. To this end, as illustrated in Fig. 3, a convolutional neural network is proposed in our framework that composes of a feature extractor (denoted as $F$) and a meta learner (denoted as $M$). Then a depth estimator (denoted as $D$) is further integrated into our network, through which domain knowledge can be incorporated. Besides, to address the second issue, the proposed framework adopts a fine-grained learning strategy that meta-learning is jointly conducted among $N-1$ meta-train domains and one meta-test domain in each iteration, by which a variety of domain shift scenarios are simultaneously exploited in each iteration. The whole meta-learning process is summarized in Algorithm 1 and the details are as follows:

\subsubsection{Meta-Train.} We sample batches in every meta-train domain $D_{trn}$, denoted as $\mathcal{\widehat{T}}_{i}$ $(i=1,...,N-1)$, and we conduct the cross-entropy classification based on the binary class labels in each meta-train domain as follows:

\begin{equation}
\small
\begin{split}
&\mathcal{L}_{Cls(\mathcal{\widehat{T}}_{i})}(\theta_F, \theta_M) \\
&= \sum\limits_{(x,y)\sim\mathcal{\widehat{T}}_{i}}ylogM(F(x))+(1-y)log(1-M(F(x)))
\end{split}
\end{equation} 
where $\theta_F$ and $\theta_M$ are the parameters of the feature extractor and the meta learner. In each meta-train domain, We can thus search the learning direction by calculating gradient of meta learner w.r.t the loss ($\nabla_{\theta_M}\mathcal{L}_{Cls(\mathcal{\widehat{T}}_{i})}(\theta_F, \theta_M)$). The updated meta learner can be calculated as ${\theta_{M_{i}}}'=\theta_M-\alpha\nabla_{\theta_M}\mathcal{L}_{Cls(\mathcal{\widehat{T}}_{i})}(\theta_F, \theta_M)$. In the meantime, we incorporate face depth maps as the domain knowledge to regularize the above learning process of the feature extractor as follows:

\begin{equation}
\small
\begin{split}
\mathcal{L}_{Dep(\mathcal{\widehat{T}}_{i})}(\theta_F, \theta_D)= \sum_{(x,I)\sim\mathcal{\widehat{T}}_{i}}\|D(F(x))-I \|^2
\end{split}
\end{equation} 
where $\theta_D$ is the parameter of the depth estimator and $I$ are the pre-calculated face depth maps for input face images. We use the state-of-the-art dense face alignment network named PRNet~\cite{2018ECCVprnet} to estimate depth maps of real faces, which serve as the supervision for the real faces. Attacks are assumed to have no face depth so that depth maps of all zeros are set as the supervision for fake faces.

\subsubsection{Meta-Test.} Moreover, we sample batch in the one remaining meta-test domain $D_{val}$, denoted as $\mathcal{\widetilde{T}}$. By adopting fine-grained learning strategy, we encourage our face anti-spoofing model trained on every meta-train domain can simultaneously perform well on the disjoint meta-test domain so that our model can be trained to generalize well to unseen attacks of various scenarios. Thus, multiple cross-entropy classifications are jointly conducted over all the updated meta learners:
\begin{equation}
\small
\begin{split}
&\sum\limits_{i=1}^{N-1}\mathcal{L}_{Cls(\mathcal{\widetilde{T}})}(\theta_F,{\theta_{M_{i}}}')=\\
&\sum\limits_{i=1}^{N-1}\sum\limits_{{(x,y)\sim\mathcal{\widetilde{T}}}}ylog{M_{i}}'(F(x))+(1-y)log(1-{M_{i}}'(F(x)))
\end{split}
\end{equation}
The domain knowledge is also incorporated like meta-train:
\begin{equation}
\small
\begin{split}
\mathcal{L}_{Dep(\mathcal{\widetilde{T}})}(\theta_F, \theta_D)= \sum_{(x,I)\sim\mathcal{\widetilde{T}}}\|D(F(x))-I \|^2
\end{split}
\end{equation}

\begin{algorithm}[t]
	\footnotesize
	\caption{\footnotesize{Regularized Fine-grained Meta Face Anti-spoofing}}
	\label{alg:A}
	\begin{algorithmic}[1]
		
		\REQUIRE~~\\
		\textbf{Input:} $N$ source domains $D=[D_1,D_2,...,D_N]$,  \\
		\textbf{Initialization:} Model parameters $\theta_F, \theta_D, \theta_M$. Hyperparameters $\alpha, \beta$
		\WHILE {not done}
		\STATE Randomly select $(N-1)$ source domains in $D$ as $D_{trn}$, and the remaining one as $D_{val}$\\
		\STATE\textbf{Meta-train}: Sampling batch in each domain in $D_{trn}$ as $\mathcal{\widehat{T}}_{i}$  $(i=1,...,N-1)$  
		\FOR{each $\mathcal{\widehat{T}}_{i}$}
		\STATE $  \mathcal{L}_{Cls(\mathcal{\widehat{T}}_{i})}(\theta_F, \theta_M)= \sum\limits_{(x,y)\sim\mathcal{\widehat{T}}_{i}}ylogM(F(x))+(1-y)log(1-M(F(x)))$
		\STATE ${\theta_{M_{i}}}'=\theta_M-\alpha\nabla_{\theta_M}\mathcal{L}_{Cls(\mathcal{\widehat{T}}_{i})}(\theta_F, \theta_M)$
		\STATE $ \mathcal{L}_{Dep(\mathcal{\widehat{T}}_{i})}(\theta_F, \theta_D)= \sum_{(x,I)\sim\mathcal{\widehat{T}}_{i}}\|D(F(x))-I \|^2$
		\ENDFOR
		
		\STATE\textbf{Meta-test}: Sampling batch in $D_{val}$ as $\mathcal{\widetilde{T}}$ 
		
		\STATE$\normalsize{\sum\limits_{i=1}^{N-1}\mathcal{L}_{Cls(\mathcal{\widetilde{T}})}(\theta_F,{\theta_{M_{i}}}')=}
		\sum\limits_{i=1}^{N-1}\sum\limits_{{(x,y)\sim\mathcal{\widetilde{T}}}}ylog{M_{i}}'(F(x))+(1-y)log(1-{M_{i}}'(F(x)))$
		\STATE $ \mathcal{L}_{Dep(\mathcal{\widetilde{T}})}(\theta_F, \theta_D)= \sum_{(x,I)\sim\mathcal{\widetilde{T}}}\|D(F(x))-I \|^2$
		\STATE\textbf{Meta-optimization}:
		\STATE$\theta_M \leftarrow \theta_M - \beta\nabla_{\theta_M}(\sum\limits_{i=1}^{N-1}(\mathcal{L}_{Cls(\mathcal{\widehat{T}}_{i})}(\theta_F, \theta_M)+\mathcal{L}_{Cls(\mathcal{\widetilde{T}})}(\theta_F,{\theta_{M_{i}}}')))$
		\STATE $\theta_F \leftarrow \theta_F - \beta\nabla_{\theta_F}(   \mathcal{L}_{Dep(\mathcal{\widetilde{T}})}(\theta_F, \theta_D) + \sum\limits_{i=1}^{N-1}( \mathcal{L}_{Cls(\mathcal{\widehat{T}}_{i})}(\theta_F, \theta_M) + \mathcal{L}_{Dep(\mathcal{\widehat{T}}_{i})}(\theta_F, \theta_D) +   \mathcal{L}_{Cls(\mathcal{\widetilde{T}})}(\theta_F,{\theta_{M_{i}}}') )   )$
		\STATE $ \theta_D \leftarrow \theta_D - \beta\nabla_{\theta_D}( \mathcal{L}_{Dep(\mathcal{\widetilde{T}})}(\theta_F, \theta_D) +
		\sum\limits_{i=1}^{N-1}( \mathcal{L}_{Dep(\mathcal{\widehat{T}}_{i})}(\theta_F, \theta_D)  ) ) $
		
		\ENDWHILE
		\RETURN Model parameters $\theta_F, \theta_D, \theta_M$
	\end{algorithmic}
\end{algorithm}

\subsubsection{Meta-Optimization.}

To summarize all the learning information in the meta-train and meta-test for optimization, we jointly train the three modules in our network as follows:
\begin{equation}
\small
\begin{split}
&\theta_M \leftarrow \theta_M - \beta\nabla_{\theta_M}(\sum\limits_{i=1}^{N-1}(\mathcal{L}_{Cls(\mathcal{\widehat{T}}_{i})}(\theta_F, \theta_M)+\mathcal{L}_{Cls(\mathcal{\widetilde{T}})}(\theta_F,{\theta_{M_{i}}}')))
\end{split}
\end{equation}
 
\begin{equation}
\small
\begin{split}
&\theta_F \leftarrow \theta_F - \beta\nabla_{\theta_F}(   \mathcal{L}_{Dep(\mathcal{\widetilde{T}})}(\theta_F, \theta_D) + \sum\limits_{i=1}^{N-1}( \mathcal{L}_{Cls(\mathcal{\widehat{T}}_{i})}(\theta_F, \theta_M) \\
&+ \mathcal{L}_{Dep(\mathcal{\widehat{T}}_{i})}(\theta_F, \theta_D) +   \mathcal{L}_{Cls(\mathcal{\widetilde{T}})}(\theta_F,{\theta_{M_{i}}}') )   )
\end{split}
\end{equation}

\begin{equation}
\small
\begin{split}
\theta_D \leftarrow \theta_D - \beta\nabla_{\theta_D}( \mathcal{L}_{Dep(\mathcal{\widetilde{T}})}(\theta_F, \theta_D) +
\sum\limits_{i=1}^{N-1}( \mathcal{L}_{Dep(\mathcal{\widehat{T}}_{i})}(\theta_F, \theta_D)  ) )
\end{split}
\end{equation}
Note that in (6), regression losses of depth estimation provides auxiliary supervision in the optimization of feature extractor. This can regularize the feature learning process of the feature extractor. In this way, the classifications in (1) and (3) within the meta learner are restrictively conducted in the feature space regularized by the auxiliary supervision of domain knowledge. This makes meta-train and meta-test focus on better-generalized learning directions.

\subsubsection{Analysis.} This section provides more detailed analysis on the proposed method. The objective of (5) in the meta-optimization is as follows (omitting $\theta_F$ for simplicity): 
\begin{equation}
\small
\begin{split}
\min_{\theta_M}\sum\limits_{i=1}^{N-1}(\mathcal{L}_{Cls(\mathcal{\widehat{T}}_{i})}(\theta_M)+\mathcal{L}_{Cls(\mathcal{\widetilde{T}})}({\theta_{M_{i}}}'))
\end{split}
\end{equation}
We do the first-order Taylor expansion on the second term as follows:
\begin{equation}
\small
\begin{split}
&\mathcal{L}_{Cls(\mathcal{\widetilde{T}})}({\theta_{M_{i}}}')=\mathcal{L}_{Cls(\mathcal{\widetilde{T}})}(\theta_M-\alpha\nabla_{\theta_M}\mathcal{L}_{Cls(\mathcal{\widehat{T}}_{i})}(\theta_M)) =\\
&\mathcal{L}_{Cls(\mathcal{\widetilde{T}})}(\theta_M)+\nabla_{\theta_M}\mathcal{L}_{Cls(\mathcal{\widetilde{T}})}(\theta_M)^{T}(-\alpha\nabla_{\theta_M}\mathcal{L}_{Cls(\mathcal{\widehat{T}}_{i})}(\theta_M))
\end{split}
\end{equation}
and the objective becomes:
\begin{equation}
\small
\begin{split}
&\min_{\theta_M}  \sum\limits_{i=1}^{N-1}(\mathcal{L}_{Cls(\mathcal{\widehat{T}}_{i})}(\theta_M)+\mathcal{L}_{Cls(\mathcal{\widetilde{T}})}(\theta_M)\\
&-\alpha (\nabla_{\theta_M}\mathcal{L}_{Cls(\mathcal{\widehat{T}}_{i})}(\theta_M)^{T} \cdotp \nabla_{\theta_M}\mathcal{L}_{Cls(\mathcal{\widetilde{T}})}(\theta_M)))
\end{split}
\end{equation}
The above objective shows that meta-optimization finds the generalized learning direction in the meta learner through: 1) minimizing losses in all meta-train and meta-test domains 2) meanwhile coordinating the learning directions (gradients information) between meta-train and meta-test so that the optimization can be conducted without overfitting to a single domain. It should be noted that there are two major differences compared to vanilla meta-learning for DG: 1) the above objective is conducted in feature space regularized by the domain knowledge supervision instead of in instance space~\cite{DaLi2017AAAI}. This makes both meta-train and meta-test focus on better-generalized learning directions and thus their learning directions are more likely to be coordinated in the task of face anti-spoofing (in the above third term). 2) vanilla meta-learning for DG~\cite{DaLi2017AAAI} is simply conducted between one aggregated meta-train domain and one aggregated meta-test domain in each iteration. Comparatively, the above objective is simultaneously conducted between multiple ($N-1$) pairs of meta-train and meta-test domains in each iteration. This adopts a fine-grained learning strategy that meta-learning is simultaneously conducted in a variety of domain shift scenarios in each iteration. Thus our face anti-spoofing model can be trained to generalize well to unseen attacks of various scenarios in each iteration.

\section{Experiments}

\subsection{Datasets}

The evaluation of our method is conducted on four public face anti-spoofing datasets that contain both print and video replay attacks: Oulu-NPU~\cite{2017FGoulu} (O for short), CASIA-MFSD~\cite{2012ICBcasia} (C for short), Idiap Replay-Attack~\cite{2012BIOSIGidiap} (I for short), and MSU-MFSD~\cite{2015TIFSida} (M for short). Table 1 in the supplementary material\footnote{Codes are available at https://github.com/rshaojimmy/AAAI2020-RFMetaFAS} shows the variations in these four datasets. Figure 1 in the supplementary material shows some samples of the genuine faces and attacks. Table 1 and Fig. 1 in supplementary material show that compared to the seen training data, attacks from unseen materials, illumination, background, resolution and so on cause significant domain shifts among these datasets.

\subsection{Experimental Setting}

Following the setting in~\cite{Shao_2019_CVPR}, one dataset is treated as one domain in our experiment. We randomly select three datasets in four as source domains where domain generalization is conducted. The left one is the unseen domain for testing, which is unavailable in the training process. Half Total Error Rate (HTER)~\cite{2004HTER} (half of the summation of false acceptance rate and false rejection rate) and Area Under Curve (AUC) are used as the evaluation metrics in our experiments.

\begin{figure*}[htb]
	\begin{subfigure}{.24\textwidth}
		\centering
		\includegraphics[width=\textwidth]{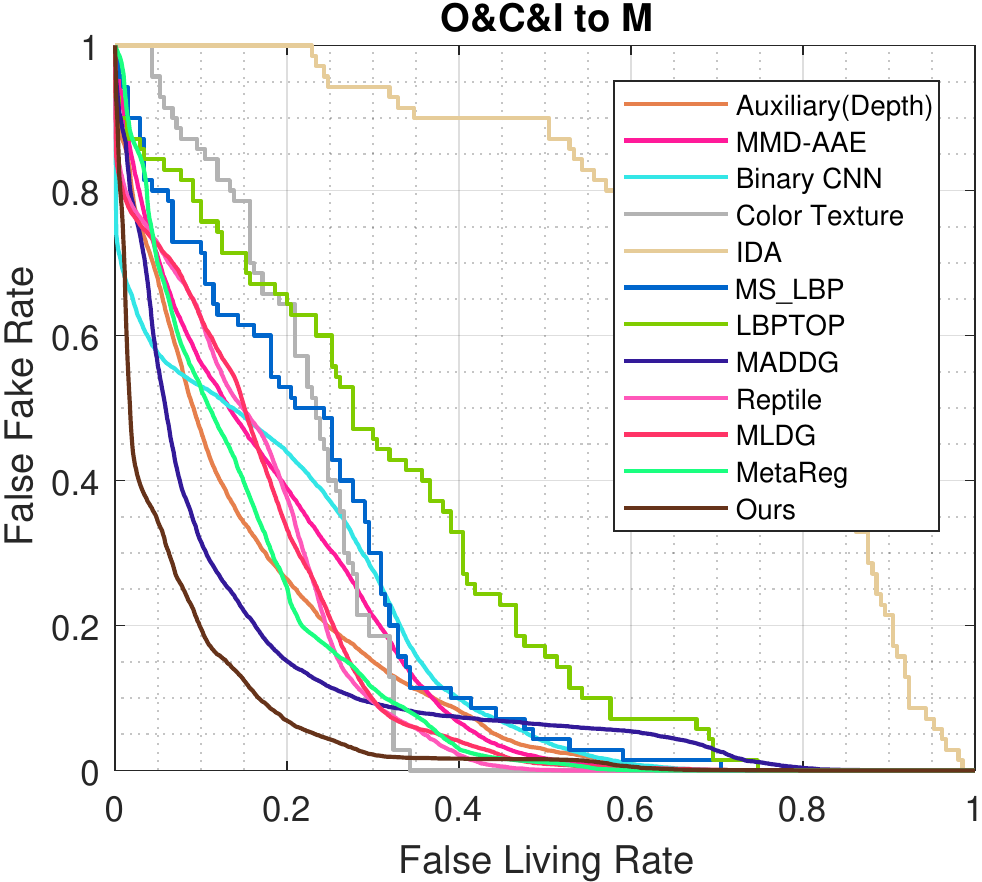}
	\end{subfigure}
	\begin{subfigure}{.24\textwidth}
		\centering
		\includegraphics[width=\textwidth]{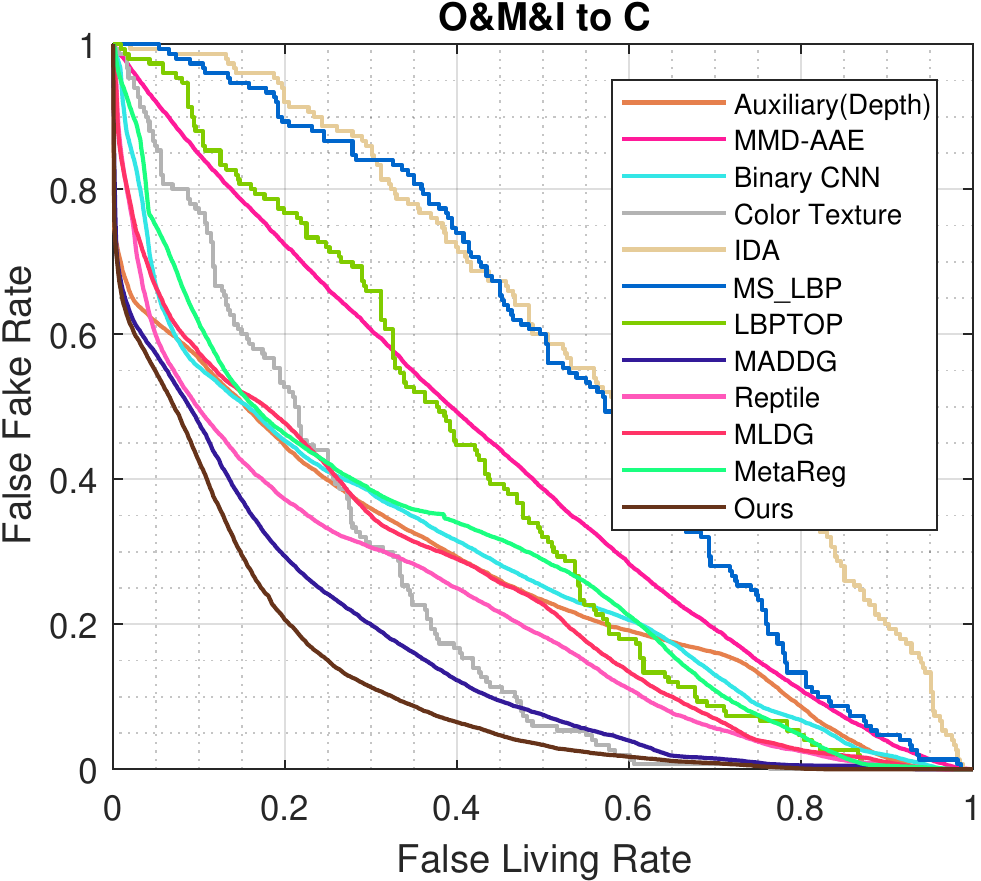}
	\end{subfigure}
	\begin{subfigure}{.24\textwidth}
		\centering
		\includegraphics[width=\textwidth]{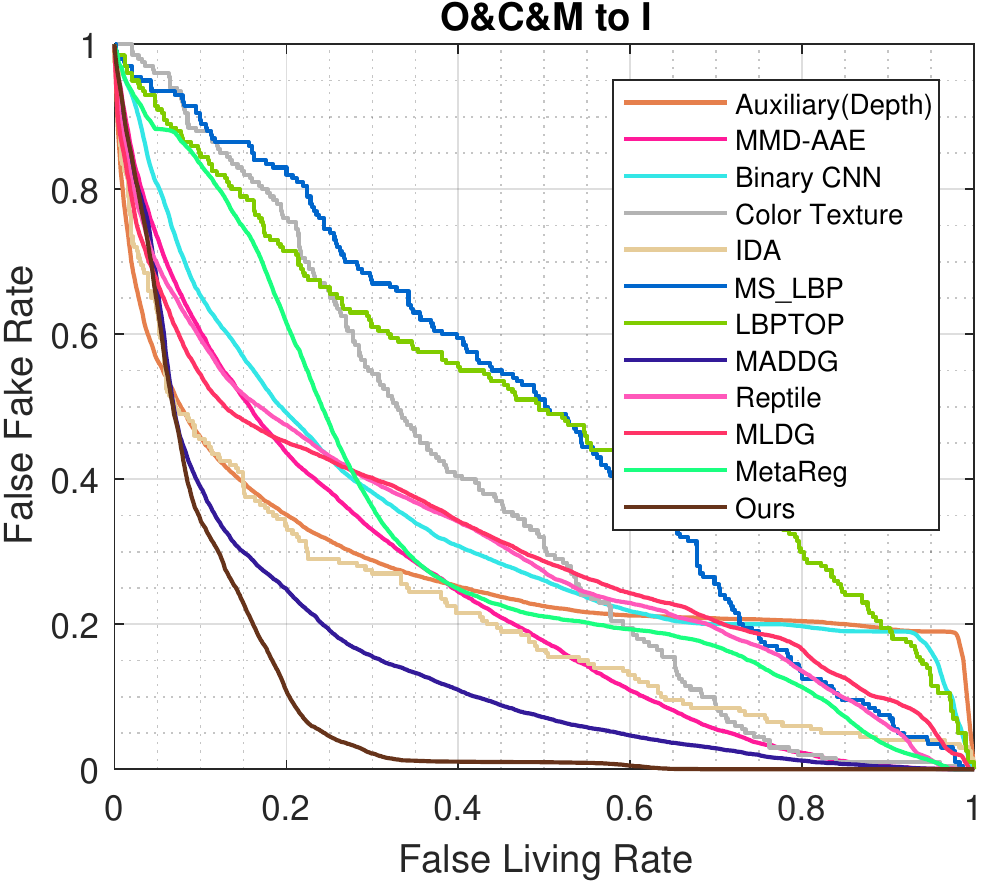}
	\end{subfigure}
	\begin{subfigure}{.24\textwidth}
		\centering
		\includegraphics[width=\textwidth]{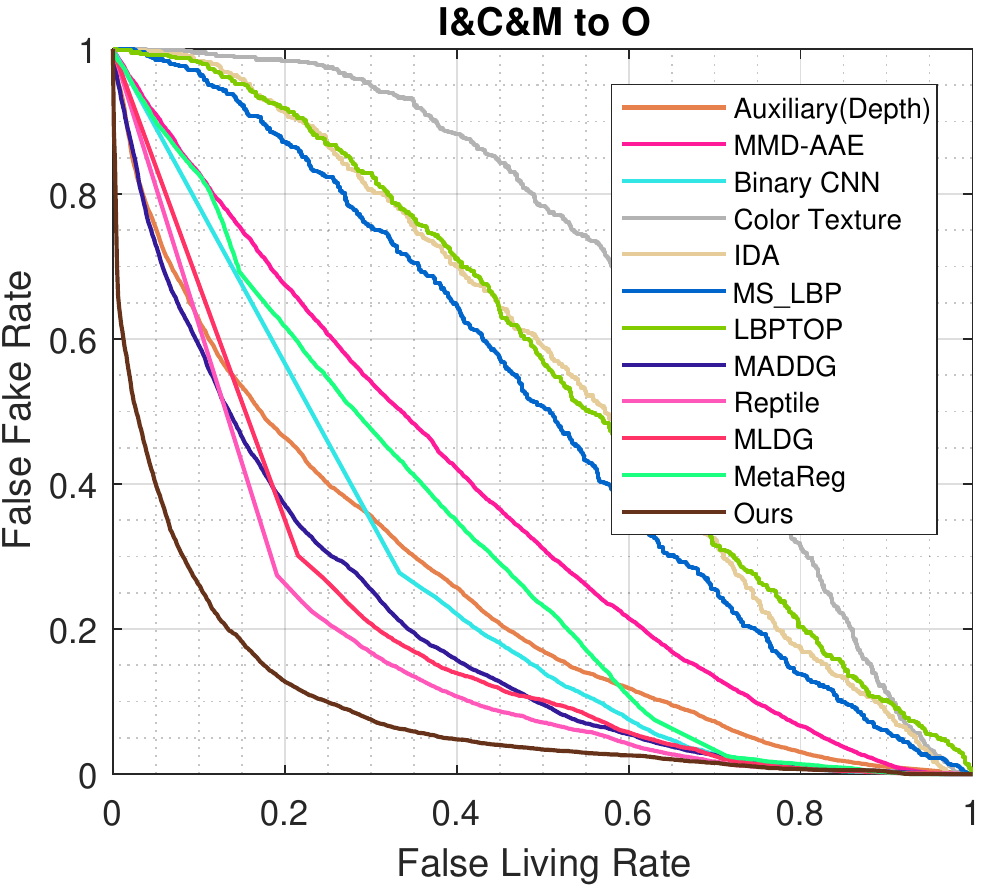}
	\end{subfigure}
	\caption{ROC curves of four testing sets for domain generalization on face anti-spoofing.}
\end{figure*}

\subsection{Implementation Details}

\textbf{Network Structure.} Our deep network is implemented on the platform of PyTorch. The detailed structure of the proposed network is illustrated in Table 2 in the supplementary material. \textbf{Training Details.} The Adam optimizer~\cite{adam} is used for the optimization. The learning rates  $\alpha, \beta$ are set as 1e-3. The batch size is 20 per domain, and thus 60 for 3 training domains totally. \textbf{Testing.} For a new testing sample $x$, its classification score $l$ is calculated for testing as follows: $l = M(F(x))$, where $F$ and $M$ are the trained feature extractor and meta learner.

\begin{table*}[!htb]
	\centering	
	\small
	\caption{Comparison to face anti-spoofing methods on four testing sets for domain generalization on face anti-spoofing.}
	\begin{tabular}{c|c|c|c|c|c|c|c|c}
		\hline
		\multirow{2}{*}{\textbf{Method}} & \multicolumn{2}{c|}{\textbf{O\&C\&I to M}} & \multicolumn{2}{c|}{\textbf{O\&M\&I to C}} 
		& \multicolumn{2}{c|}{\textbf{O\&C\&M to I}}& \multicolumn{2}{c}{\textbf{I\&C\&M to O}}\\ \cline{2-9} 		
		& HTER(\%)                                   & AUC(\%)                                   & HTER(\%)                                   & AUC(\%)                                   & HTER(\%)                                   & AUC(\%)                                   & HTER(\%)                                   & AUC(\%)                                   \\ \hline
		MS\_LBP							            &29.76&78.50	&54.28&44.98	&50.30&51.64 	&50.29&49.31\\	
		Binary CNN									&29.25 &82.87 	&34.88&71.94 	&34.47&65.88 	&29.61&77.54\\
		IDA											&66.67&27.86	&55.17&39.05	&28.35&78.25	&54.20&44.59\\
		Color Texture       						&28.09&78.47    &30.58&76.89    &40.40&62.78    &63.59&32.71\\
		LBPTOP       								&36.90&70.80    &42.60&61.05    &49.45&49.54    &53.15&44.09\\		
		Auxiliary(Depth Only)						&22.72&85.88	&33.52&73.15	&29.14&71.69	&30.17&77.61\\
		Auxiliary(All)	 							&--&--			&28.4&--	    &27.6&--	    &--&--\\
		MMD-AAE                						&27.08&83.19 	&44.59&58.29	&31.58&75.18	&40.98&63.08\\			
		MADDG           							&17.69&88.06 	&24.5&84.51	&22.19&84.99 		&27.98&80.02\cr\hline
		\textbf{Ours}               				&\textbf{13.89}&\textbf{93.98} 	&\textbf{20.27}&\textbf{88.16}	&\textbf{17.3}&\textbf{90.48}		&\textbf{16.45}&\textbf{91.16}\\\hline		
		
	\end{tabular}
\end{table*}

\begin{table*}[!htb]
	\centering	
	\small
	\caption{Comparison to meta-learning for DG methods on four testing sets for domain generalization on face anti-spoofing.}
	\begin{tabular}{c|c|c|c|c|c|c|c|c}
		\hline
		\multirow{2}{*}{\textbf{Method}} & \multicolumn{2}{c|}{\textbf{O\&C\&I to M}} & \multicolumn{2}{c|}{\textbf{O\&M\&I to C}} 
		& \multicolumn{2}{c|}{\textbf{O\&C\&M to I}}& \multicolumn{2}{c}{\textbf{I\&C\&M to O}}\\ \cline{2-9} 	 
		& HTER(\%)                                   & AUC(\%)                                   & HTER(\%)                                   & AUC(\%)                                   & HTER(\%)                                   & AUC(\%)                                   & HTER(\%)                                   & AUC(\%)                                   \\ \hline
		Reptile                						&23.64&85.06 	&30.38&78.10	&36.13&69.01	&22.88&82.22\\
		MLDG                						&23.91&84.81 	&32.75&74.51	&36.55&68.54  	&25.75&79.52\\
		MetaReg                						&21.17&86.11    &35.66&70.83    &32.28&67.48  	&37.72&68.71\\\hline			
		\textbf{Ours}               				&\textbf{13.89}&\textbf{93.98} 	&\textbf{20.27}&\textbf{88.16}	&\textbf{17.3}&\textbf{90.48}		&\textbf{16.45}&\textbf{91.16}\\\hline		
		
	\end{tabular}
\end{table*}

\subsection{Experimental Comparison}

\subsubsection{Baseline Methods.}

We compare several state-of-the-art face anti-spoofing methods as follows: \textbf{Multi-Scale LBP (MS\_LBP)}~\cite{2011IJCBmstexture} ; \textbf{Binary CNN}~\cite{2014arxivdeepfeat}; \textbf{Image Distortion Analysis (IDA)}~\cite{2015TIFSida}; \textbf{Color Texture (CT)}~\cite{2016TIFScolortxt}; \textbf{LBPTOP}~\cite{2014EJIVPlbptop}; \textbf{Auxiliary}~\cite{2018CVPRauxliary}: To fairly compare our method only using one frame information, we implement its face depth estimation component(denoted as Auxiliary(Depth Only)). We also compare its reported results (denoted as Auxiliary(All));  \textbf{MMD-AAE}~\cite{2018CVPRdgadv}; and \textbf{MADDG}~\cite{Shao_2019_CVPR}. Moreover, we also compare the related state-of-the-art meta-learning for DG methods in the face anti-spoofing task: \textbf{MLDG}~\cite{DaLi2017AAAI}; \textbf{Reptile}~\cite{Reptile_2018_arXiv}; and \textbf{MetaReg}~\cite{MetaReg_2018_NIPS}.

\subsubsection{Comparison Results.}

From comparison results in Table 1 and Fig. 4, it can be seen that the proposed method outperforms the state-of-the-art face anti-spoofing methods~\cite{2011IJCBmstexture,2014arxivdeepfeat,2015TIFSida,2016TIFScolortxt,2018CVPRauxliary}. This is because all these methods focus on extracting differentiation cues the only fit to attacks in the source domains. Comparatively, the proposed meta-learning for DG trains our face anti-spoofing model to generalize well in the simulated domain shift scenario. This significantly improves the generalization ability of the face anti-spoofing method. Moreover, we also compare the DG with adversarial learning methods for face anti-spoofing~\cite{2018CVPRdgadv,Shao_2019_CVPR} and our method also performs better. This is because instead of focusing on learning a domain shared feature space and training extra domain discriminators, our method just needs to train a simple network with meta-learning strategy. This realizes the DG for face anti-spoofing in a more feasible and efficient way.  

Table 2 and Fig. 4 show that compared to some state-of-the-art vanilla meta-learning for DG methods~\cite{DaLi2017AAAI,Reptile_2018_arXiv}, our method also outperforms them for the task of face anti-spoofing. This illustrates that by addressing the above two issues, the proposed meta-learning framework is more able to improve the generalization ability for the task of face anti-spoofing.

\subsection{Ablation Study}

\subsubsection{Components Evaluation.}

\begin{figure}[htb]
	\begin{subfigure}{.23\textwidth}
		\centering
		\includegraphics[width=\textwidth]{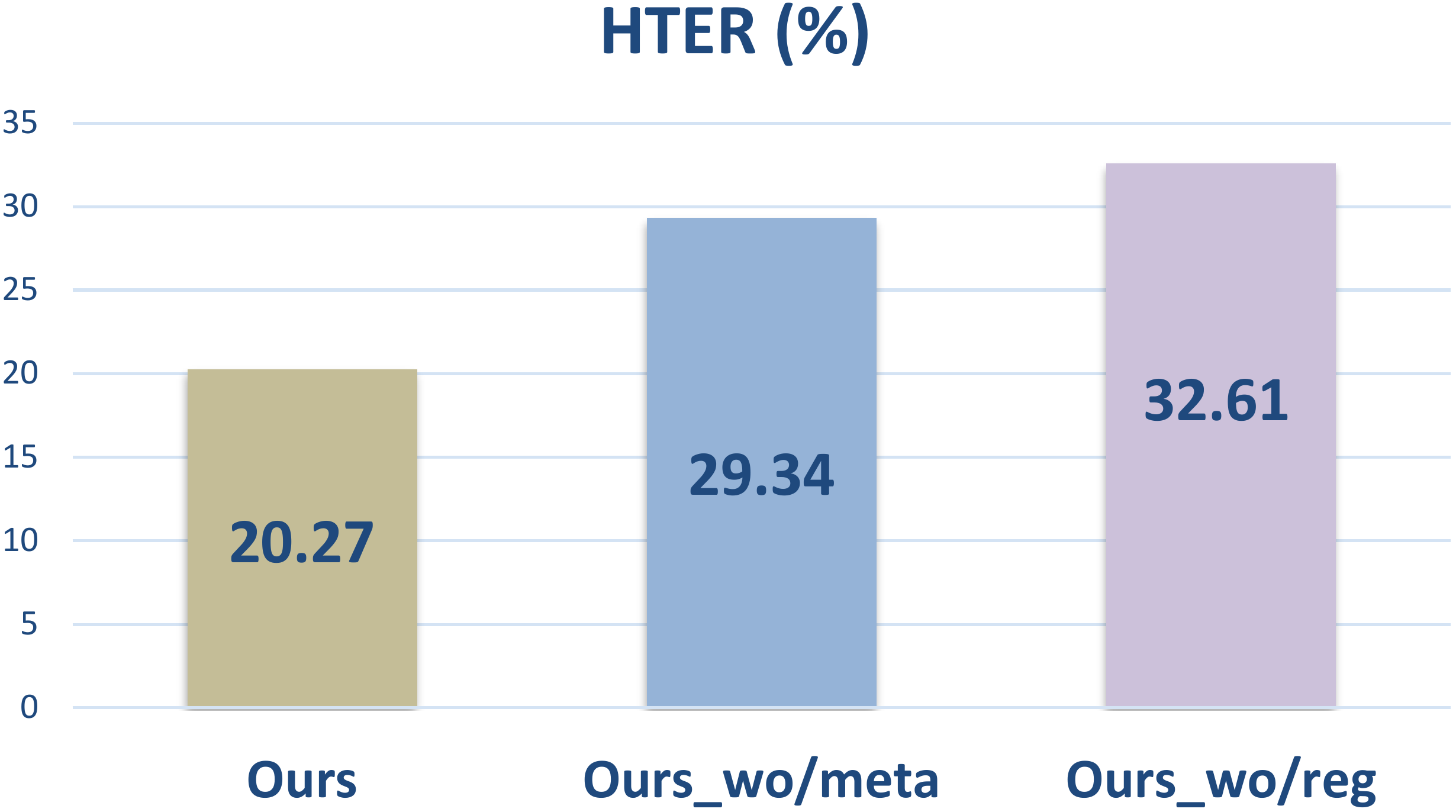}
	\end{subfigure}
	\begin{subfigure}{.23\textwidth}
		\centering
		\includegraphics[width=\textwidth]{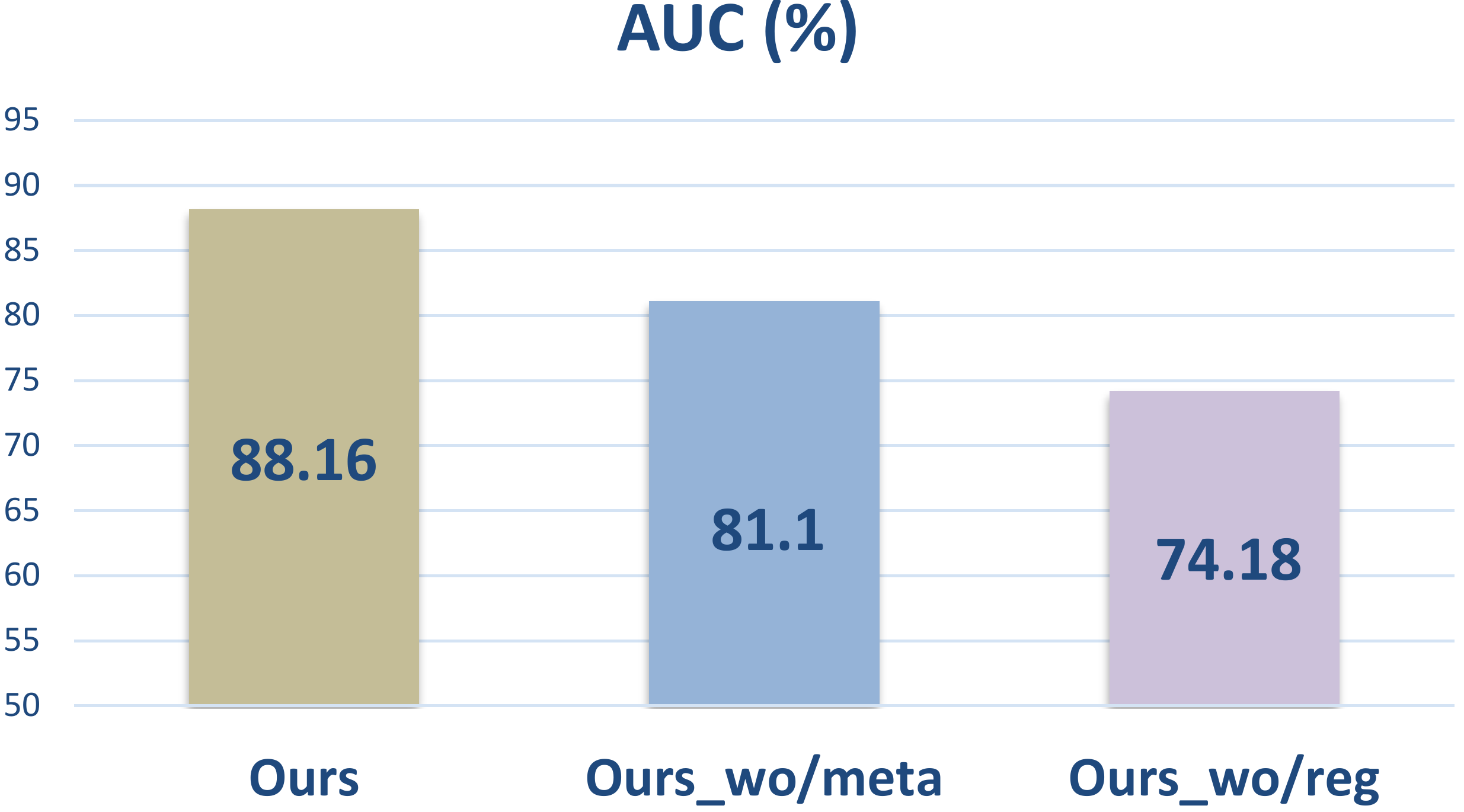}
	\end{subfigure}
	\caption{Evaluation of different components of proposed method in O\&M\&I to C set for face anti-spoofing.}
\end{figure}

\begin{table*}[!htb]
	\centering	
	\small
	\caption{Effectiveness of fine-grained learning strategy and second-order derivative information}
	\begin{tabular}{c|c|c|c|c|c|c|c|c}
		\hline
		\multirow{2}{*}{\textbf{Method}} & \multicolumn{2}{c|}{\textbf{O\&C\&I to M}} & \multicolumn{2}{c|}{\textbf{O\&M\&I to C}} 
		& \multicolumn{2}{c|}{\textbf{O\&C\&M to I}}& \multicolumn{2}{c}{\textbf{I\&C\&M to O}}\\ \cline{2-9} 	 
		& HTER(\%)                                   & AUC(\%)                                   & HTER(\%)                                   & AUC(\%)                                   & HTER(\%)                                   & AUC(\%)                                   & HTER(\%)                                   & AUC(\%)                                   \\ \hline
		Ours (Aggregation)                					  &14.54&92.87 	&24.28&85.29	&20.07&88.13  	&17.94&90.69\\		
		Ours (First-order)                		  &17.93&87.36 	&27.47&82.17	&26.24&79.32	&19.24&87.82\\\hline			
		\textbf{Ours}               				&\textbf{13.89}&\textbf{93.98} 	&\textbf{20.27}&\textbf{88.16}	&\textbf{17.3}&\textbf{90.48}		&\textbf{16.45}&\textbf{91.16}\\\hline		
		
	\end{tabular}
\end{table*}

Considering that O\&M\&I to C set has the most significant domain shift, we evaluate different components of our method in this set for an example and experimental results are shown in Fig. 5. \textbf{Ours} denotes the proposed method. \textbf{Ours\_wo/meta} denotes the proposed network without the meta-learning component. In this setting, we do not conduct the meta-learning in the meta learner part. \textbf{Ours\_wo/reg} denotes the proposed network without domain knowledge regularization. In this setting, we do not incorporate the face depth maps as the domain knowledge to regularize the meta-learning process. 

Figure 5 shows that the proposed network has degraded performance if any component is excluded. Specifically, the results of \textbf{Ours\_wo/meta} verify that the meta-learning conducted in the meta learner benefits for the generalization ability improvement. The results of \textbf{Ours\_wo/reg} show that without the regularization of domain knowledge supervision, the performance of our meta-learning for DG degrades significantly. This validates that by addressing the first issue, the proposed meta-learning framework is more able to develop a generalized face anti-spoofing model.

\begin{figure}[htb]
	\begin{center}
		\includegraphics[height=7.1cm, width=0.75\linewidth]{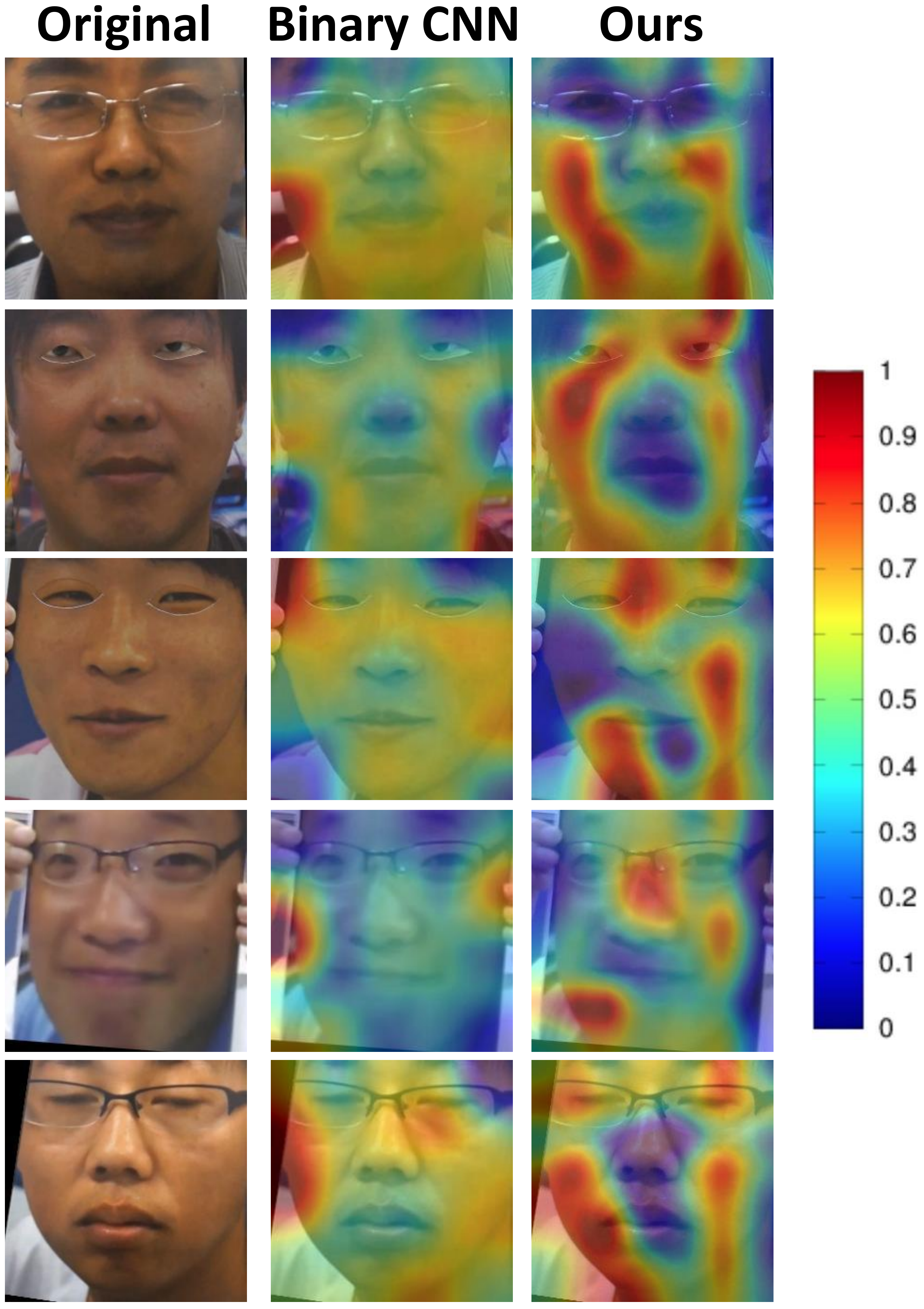}
	\end{center}
	\caption{Attention map visualization of Binary CNN and our method for testing samples of attacks in O\&M\&I to C set. (Best reviewed in colors)}
\end{figure}

\subsubsection{Effectiveness of fine-grained learning strategy and second-order derivative information.} As mentioned in the above analysis, compared to vanilla meta-learning for DG methods, our method adopts a fine-grained learning strategy which can help to develop face anti-spoofing model with the generalization ability to unseen attacks of various scenarios. To verify the effectiveness of this strategy, we conduct our method in the setting proposed in~\cite{DaLi2017AAAI}, where the proposed regularized meta-learning is only conducted between one aggregated meta-train and one aggregated meta-test domains in each training iteration. The comparison results are named as Ours (aggregation) in Table 3. Table 3 shows that our method obtains better performance than Ours (aggregation). This validates that the proposed meta-learning adopting fine-grained learning strategy is more able to improve the generalization ability for the task of face anti-spoofing. Moreover, the third term in (10) has the function of coordinating the learning of meta-train and meta-test so as to prevent the optimization process from overfitting to a single domain. This improves the generalization ability but at the same time involves the second-order derivative computation of parameters of meta learner. Some works such as Reptile~\cite{Reptile_2018_arXiv} uses a first-order approximation to decrease the computation complexity. We thus compare a method named as Ours (First-order) in Table 3 that replaces the second-order derivative computation in meta learner with the first-order approximation proposed in Reptile~\cite{Reptile_2018_arXiv}. Results show that our method performs better, which verifies that the second-order derivative information in the third term of (10) is more effective and plays a key role in the generalization ability improvement for the task of face anti-spoofing.

\subsection{Attention Map Visualization}

To provide more insights on why our method improves the generalization ability for the task of face anti-spoofing, we visualize the attention map of networks by the Global Average Pooling (GAP) method~\cite{Bolei_2016_CVPR}. Figure 6 shows some examples of visualization results for the testing samples of attacks between Binary CNN~\cite{2014arxivdeepfeat} and our method. In~\cite{2014arxivdeepfeat}, authors train a CNN only with supervision of binary class labels in the face anti-spoofing task. This makes the model focus on capturing biased differentiation cues with poor generalization ability. In the visualization of Binary CNN of Fig. 6, it can be seen that when encountering unseen testing attacks, this method pays the most attention to extracting the differentiation cues in the background (row 1-2) or on paper edges/holding fingers (row 3-5). These differentiation cues are not generalized because they will be changed if the attacks are from a new background or without clear paper edges. Comparatively, Fig. 6 shows that our method always focuses on the region of internal face for searching differentiation cues. These differentiation cues are more likely to be intrinsic and generalized for face anti-spoofing and thus the generalization ability of our method can be improved.

\section{Conclusion}

To improve the generalization ability of face anti-spoofing methods, this paper casts face anti-spoofing as a domain generalization problem, which is addressed in a new regularized fine-grained meta-learning framework. The proposed framework conducts meta-learning in the feature space regularized by the domain knowledge supervision. In this way, better-generalized learning information for face anti-spoofing can be meta-learned. Besides, a fine-grained learning strategy is adopted which enables a variety of domain shift scenarios to be simultaneously exploited for meta-learning so that our model can be trained to generalize well to unseen attacks of various scenarios. Comprehensive experimental results validate the effectiveness of the proposed method statistically and visually. 

\textbf{Acknowledgments} This project is partially supported by Hong Kong RGC GRF HKBU12200518. The work of X. Lan is partially supported by HKBU Tier 1 Start-up Grant.

\bibliographystyle{aaai} 	
\bibliography{aaai}

\section{Supplementary Material}

\subsection{Datasets}

\begin{figure*}[t]
	\begin{center}
		\includegraphics[height=2.5cm, width=1\linewidth]{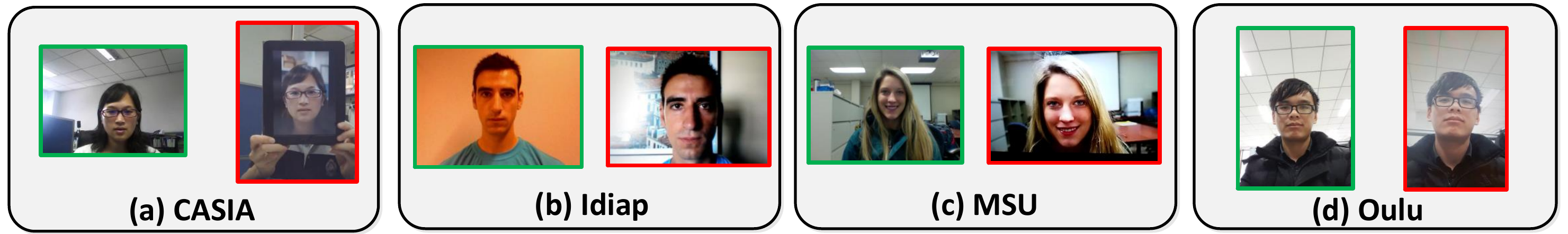}
	\end{center}
	\caption{ Sample frames from CASIA-MFSD~\cite{2012ICBcasia}, Idiap Replay-Attack~\cite{2012BIOSIGidiap}, MSU-MFSD~\cite{2015TIFSida}, and Oulu-NPU~\cite{2017FGoulu} datasets. The figures with green border represent the real faces, while the ones with red border represent the video replay attacks. From these examples, it can be seen that large cross-dataset variations due to the differences on materials, illumination, background, resolution and so on, cause significant domain shift among these datasets.}
\end{figure*}

\begin{table}[htb]
	\renewcommand{\arraystretch}{1}
	\centering	
	\scriptsize
	\caption{Comparison of four experimental datasets.}	
	\begin{tabular}{c|c|c|c|c}
		\hline\hline
		\textbf{Dataset} & \begin{tabular}[c]{@{}c@{}}\textbf{Light }\\\textbf{variation}\end{tabular} & \begin{tabular}[c]{@{}c@{}}\textbf{Complex}\\ \textbf{background}\end{tabular} & \begin{tabular}[c]{@{}c@{}}\textbf{Attack}\\ \textbf{type}\end{tabular}                                  & \begin{tabular}[c]{@{}c@{}}\textbf{Display} \\ \textbf{devices}\end{tabular}           \\ \hline
		C       & No                                                     & Yes                                                           & \begin{tabular}[c]{@{}c@{}}Printed photo\\ Cut photo\\ Replayed video\end{tabular}     & iPad                                                                 \\ \hline
		I       & Yes                                                    & Yes                                                          & \begin{tabular}[c]{@{}c@{}}Printed photo\\ Display photo\\ Replayed video\end{tabular} & \begin{tabular}[c]{@{}c@{}}iPhone 3GS \\ iPad\end{tabular}           \\ \hline
		M       & No                                                     & Yes                                                          & \begin{tabular}[c]{@{}c@{}}Printed photo\\ Replayed video\end{tabular}                 & \begin{tabular}[c]{@{}c@{}}iPad Air\\ iPhone 5S\end{tabular}         \\ \hline
		O       & Yes                                                    & No                                                           & \begin{tabular}[c]{@{}c@{}}Printed photo\\ Display photo\\ Replayed video\end{tabular} & \begin{tabular}[c]{@{}c@{}}Dell 1905FP\\ Macbook Retina\end{tabular} \\ \hline\hline
	\end{tabular}
\end{table}

The evaluation of our method is conducted on four public face anti-spoofing datasets that contain both print and video replay attacks: Oulu-NPU~\cite{2017FGoulu} (O for short), CASIA-MFSD~\cite{2012ICBcasia} (C for short), Idiap Replay-Attack~\cite{2012BIOSIGidiap} (I for short), and MSU-MFSD~\cite{2015TIFSida} (M for short). From Table 4 and Fig. 7, it can be seen that many kinds of variations, due to the differences on materials, illumination, background, resolution and so on, exist across these four datasets. Therefore, significant domain shift exists among these datasets.

\subsection{Network Structure}

The detailed structure of the proposed network is illustrated in Table 5. To be specific, each convolutional layer in the feature extractor, meta learner and depth estimator is followed by a batch normalization layer and a rectified linear unit (ReLU) activation function, and all convolutional kernel size is 3$\times$3. The size of input image is $256\times256\times6$, where we extract the RGB and HSV channels of each input image. Inspired by the residual network~\cite{2016CVPRresnet}, we use a short-cut connection, which is concatenating the responses of pool1-1, pool1-2 and pool1-3, and sending them to conv3-1 for depth estimation. This operation helps to ease the training procedure.

\begin{table*}[!htb]
	\renewcommand{\arraystretch}{1.2}
	\centering
	\normalsize
	\caption{The structure details of all components of the proposed network. }
	
	\begin{tabular}{ccc||ccc||ccc}
		
		\hline \hline
		\multicolumn{3}{c||}{\multirow{2}{*}{\begin{tabular}[c]{@{}c@{}}
					\textbf{Feature Extractor}\\ Layer \quad  Chan./Stri. \quad  Out.Size\end{tabular}}} 
		
		
		& \multicolumn{3}{c||}{\multirow{2}{*}{\begin{tabular}[c]{@{}c@{}}
					\textbf{Meta Learner}\\ Layer \quad  Chan./Stri.  \quad Outp.Size\end{tabular}}} 
		
		& \multicolumn{3}{c}{\multirow{2}{*}{\begin{tabular}[c]{@{}c@{}}
					\textbf{Depth Estimator}\\ Layer \quad  Chan./Stri.  \quad Outp.Size\end{tabular}}} \\
		
		\multicolumn{3}{c||}{}&\multicolumn{3}{c||}{}&\multicolumn{3}{c}{}\\ \hline
		\multicolumn{3}{c||}{\begin{tabular}[c]{@{}c@{}}Input\\image\end{tabular}}& \multicolumn{3}{c||}{\begin{tabular}[c]{@{}c@{}}Input\\ pool1-3\end{tabular}}& \multicolumn{3}{c}{\begin{tabular}[c]{@{}c@{}}Input\\pool1-1+pool1-2+pool1-3\end{tabular}} \\\hline
		conv1-1 \qquad& 64/1  \qquad& 256  &   conv2-1 \qquad& 128/1 \qquad&32  & conv3-1\qquad& 128/1 & 32  \\
		conv1-2 \qquad& 128/1 \qquad& 256  &   pool2-1 \qquad& -/2   \qquad&16  & conv3-2 \qquad&64/1 & 32   \\
		conv1-3 \qquad& 196/1 \qquad& 256  &   conv2-2 \qquad& 256/1 \qquad&16  & conv3-3 \qquad&1/1& 32    \\
		conv1-4 \qquad& 128/1 \qquad& 256  &   pool2-2 \qquad& -/2   \qquad&8   &  &  &    \\
		pool1-1 \qquad& -/2	  \qquad& 128  &   conv2-2 \qquad& 512/1 \qquad&8   &  & &    \\
		conv1-5 \qquad& 128/1 \qquad& 128  &           Average pooling     &  &   \\
		conv1-6 \qquad& 196/1 \qquad& 128  &    fc2-1   \qquad& 1/1	\qquad&1   &  &   \\
		conv1-7 \qquad& 128/1 \qquad& 128  &    	& &  &  &   \\
		pool1-2 \qquad& -/2	  \qquad& 64   &  	  	&   &  &  & &    \\
		conv1-8 \qquad& 128/1 \qquad& 64   &   	   	&    &   &  &   \\
		conv1-9 \qquad& 196/1 \qquad& 64   &   	  	&    &   &  &   \\
		conv1-10 \qquad& 128/1\qquad& 64   &      	&    &   &  &   \\		
		pool1-3 \qquad& -/2	  \qquad& 32   &    	&   &  &  & &    \\		
		\hline \hline

	\end{tabular}
\end{table*}

\end{document}